%% file: root.tex
\documentclass[letterpaper, 10 pt, conference]{ieeeconf}  

\input{setup}

\title{\LARGE \bf
FOGMACHINE - Leveraging Discrete-Event Simulation and Scene Graphs for Modeling Hierarchical, Interconnected Environments under Partial Observations from Mobile Agents
}

\author{Lars Ohnemus$^{1,2}$, Nils Hantke$^{1}$, Max Weißer$^{1}$, and Kai Furmans$^{1}$
\thanks{$^{1}$Authors are with the Institute for Material Handling and Logistics,
        Karlsruhe Institute of Technology, Karlsruhe, Germany}%
\thanks{$^{2}$Corresponding author: lars.ohnemus\@kit.edu}%
}

\begin{document}

\newcommand\submittedtext{%
  \footnotesize This work has been submitted to the IEEE for possible publication. Copyright may be transferred without notice, after which this version may no longer be accessible.}

\newcommand\submittednotice{%
\begin{tikzpicture}[remember picture,overlay]
\node[anchor=south,yshift=10pt] at (current page.south) {\fbox{\parbox{\dimexpr0.65\textwidth-\fboxsep-\fboxrule\relax}{\footnotesize\submittedtext}}};
\end{tikzpicture}%
}

\maketitle
\setcounter{footnote}{2}
\thispagestyle{empty}
\pagestyle{empty}
\submittednotice

\input{02_figures/fig01_overview}

\begin{abstract}
\input{01_sections/00_abstract}
\end{abstract}

\input{01_sections/01_introduction}
\input{01_sections/02_related_work}

\input{01_sections/03_methods}
\input{01_sections/04_application}

\input{01_sections/05_conclusion}
\input{01_sections/98_contributions}
\input{01_sections/99_acknowledgements}

\printbibliography

\end{document}

%% file: setup.tex
\IEEEoverridecommandlockouts                              

\usepackage{lipsum}
\usepackage{amsmath}
\usepackage{amssymb}
\usepackage{biblatex}
\usepackage{acronym}
\usepackage{graphicx}
\usepackage{subcaption} 
\usepackage{pgfplotstable} 
\usepackage{booktabs}      
\usepackage{multirow}
\usepackage{tabularx}
\usepackage{array} 
\usepackage{csvsimple}
\usepackage{siunitx}       
\usepackage{hyperref}
\usepackage{xfp} 
\usepackage{layouts} 
\usepackage{float} 

\definecolor{palette1}{RGB}{102,194,165}
\definecolor{palette2}{RGB}{252,141,98}
\definecolor{palette3}{RGB}{141,160,203}
\definecolor{palette4}{RGB}{231,138,195}
\definecolor{palette5}{RGB}{166,216,84}
\definecolor{palette6}{RGB}{255,217,47}
\definecolor{palette7}{RGB}{229,196,148}
\definecolor{palette8}{RGB}{179,179,179}
\definecolor{palette9}{RGB}{102,194,165}
\definecolor{palette10}{RGB}{252,141,98}

\definecolor{bgDarkEmph}{RGB}{67,80,99}

\definecolor{agent}{RGB}{43, 189, 215}
\definecolor{place}{RGB}{120, 167, 213}
\definecolor{object}{RGB}{229, 107, 109}
\definecolor{poi}{RGB}{247, 187, 97}
\definecolor{processesBackground}{RGB}{248, 216, 134}
\definecolor{agentsBackground}{RGB}{239, 195, 202}
\definecolor{trueSGBackground}{RGB}{203, 241, 253}
\definecolor{obsSGBackground}{RGB}{228, 251, 184}

\usetikzlibrary{positioning}
\usetikzlibrary{shapes,arrows}
\usetikzlibrary{calc}
\usetikzlibrary{fit,backgrounds}

\newcommand*\circnum[1]{%
  \tikz[baseline=(char.base)]{
    \node[
      circle,
      fill=bgDarkEmph,     
      text=white,        
      inner sep=0.1pt,
      minimum size=1.1em,
      line width=0pt,    
      font=\footnotesize
    ] (char) {#1};
  }%
}

\input{02_figures/00_tikz_styles}

\newcommand{\numpercent}[2][]{%
  \num[#1]{\fpeval{100*(#2)}}
}

\newcommand{\numWithUncertaintyFromCSV}[4][]{%

  \pgfplotstableread[col sep=comma]{#2}\datatable
  \pgfplotstableset{every column/.append style={string type}}

  \pgfplotstablegetelem{0}{#3Mean}\of{\datatable}%
  \pgfmathsetmacro{\val}{\pgfplotsretval}
  \pgfplotstablegetelem{0}{#3Std}\of{\datatable}%
  \pgfmathsetmacro{\stdval}{\pgfplotsretval}

  $\num[#1]{\val} \,\pm\, \num[#1]{\stdval}\,\si{#4}$%
}

\newcommand{\fogmachine}{\emph{FOGMACHINE}\,}

\acrodef{poi}[PoI]{Point of Interest}
\acrodef{sg}[SG]{Scene Graph}
\acrodef{dsg}[DSG]{Dynamic Scene Graph}
\acrodef{des}[DES]{Discrete-Event Simulation}
\acrodef{nhpp}[NHPP]{Non-Homogeneous Poisson Process}
\acrodef{adr}[ADR]{Autonomous Delivery Robot}
\acrodef{osm}[OSM]{Open Street Map}
\acrodef{pomdp}[POMDP]{Partially Observable Markov Decision Process}

\newcommand{\metricUp}{↑}
\newcommand{\metricDown}{↓}

\usepackage[capitalize,nameinlink]{cleveref}

\crefname{figure}{Fig.}{Figs.}
\Crefname{figure}{Fig.}{Figs.}
\crefname{section}{Sec.}{Secs.}
\Crefname{section}{Sec.}{Secs.}
\crefname{subsection}{Sec.}{Secs.}
\Crefname{subsection}{Sec.}{Secs.}

%% file: 02_figures/00_tikz_styles.tex
\usetikzlibrary{shapes.geometric, arrows, positioning}
\usetikzlibrary{shadings}

\pgfdeclarelayer{backgroundGlob} 
\pgfdeclarelayer{background} 
\pgfsetlayers{backgroundGlob,background,main}

\tikzset{
  font=\footnotesize\sffamily,  
  >=latex,
}

\tikzset{
  arrowlabel/.style={font=\scriptsize\sffamily},   
  boxfont/.style={font=\scriptsize\sffamily},   
  heading/.style={font=\footnotesize\bfseries\sffamily, inner sep=1pt}, 
}

\tikzset{
  emphbox/.style={
    fill=lightgray!30,
    rounded corners=2pt,
    inner sep=3pt,
    boxfont
  }
}

\newcommand{\fogoverlay}[6]{%
  \begin{scope}[even odd rule]
    \clip (#1.south west) rectangle (#1.north east);
    \fill[#5,opacity=#6]
      (#1.south west) rectangle (#1.north east)
      \foreach \obs in {#2} {(\obs) circle (#4)};
  \end{scope}

  \foreach \obs in {#2} {
    \foreach \i in {0,...,19} {
      \pgfmathsetmacro{\t}{\i/20}%
      \pgfmathsetmacro{\rin}{(#3 + \t*(#4-#3))}%
      \pgfmathsetmacro{\rout}{(#3 + (\t + 1/20)*(#4-#3))}%
      \pgfmathsetmacro{\opacity}{\t*#6}%

      \begin{scope}[even odd rule]
        \clip (#1.south west) rectangle (#1.north east);
        \fill[#5,opacity=\opacity]
          (\obs) circle[radius=\rout pt]
          (\obs) circle[radius=\rin pt];
      \end{scope}
    }
  }
}

\tikzset{
  expl/.style={
    draw,
    fill=gray!5,
    rounded corners=2pt,
    align=left,
    inner sep=2pt,
    font=\scriptsize\sffamily,
  },
  ann/.style={
    fill=white,
    rounded corners=2pt,
    align=left,
    inner sep=2pt,
    font=\scriptsize\sffamily,
  }
}

\tikzset{
  spanxy/.code 2 args={%
    \pgfkeysalso{fit={(#1.west |- #2.north) (#1.east |- #2.south)}, inner sep=0pt}%
  }
}

\tikzset{
  system/.style={
    draw,
    very thick,
    rounded corners,
    inner sep=1em,
    fill=gray!5,
  }
}

\tikzset{
  sgbox/.style={
    draw,
    thick,
    rectangle,
    rounded corners,
    align=center,
    minimum width=3cm,
    minimum height=1.5cm,
    fill=blue!10,
    boxfont
  },
  sgtrue/.style={sgbox, fill=trueSGBackground!50},
  sgobs/.style={sgbox, fill=obsSGBackground!50},
}

\tikzset{
  process/.style={
    draw,
    thick,
    rectangle,
    rounded corners,
    minimum width=2.5cm,
    minimum height=1cm,
    fill=processesBackground!40,
    align=center,
    boxfont
  },
  agent/.style={
    process,
    fill=agentsBackground!40,
  },
}

\tikzset{
  flowbox/.style={
    draw,
    thick,
    rectangle,
    minimum width=5mm,
    inner sep=2pt,
    align=center,
    fill=white,
    boxfont
  },
  flowchute/.style={
    draw,
    thick,
    trapezium,
    trapezium angle=70,
    minimum width=5mm,   
    minimum height=4mm,
    align=center,
    inner sep=0,
    fill=white,
    boxfont
  },
  flowdashed/.style={
    flowbox,
    dashed,
    fill=gray!10,
  },
}

\tikzset{
  place/.style={
    circle,
    draw=place!50!black,
    fill=place,
    minimum size=2mm,
    inner sep=0pt,
  },
  poi/.style={
    circle,
    fill=poi,
    draw=poi!50!black,
    minimum size=2mm,
    inner sep=0pt,
  },
  object/.style={
    circle,
    fill=object,
    draw=object!50!black,
    minimum width=2mm,
    minimum height=2mm,
    inner sep=0pt,
  },
  relation/.style={
    semithick,
  },
  relation-undirected/.style={
    relation,
    -,
  },
  relation-directed/.style={
    relation,
    ->,
  },
}

\tikzset{
  mobileagent/.style={
    rectangle,
    rounded corners=2pt,
    draw=agent!50!black,
    fill=agent,
    minimum width=2.5mm,
    minimum height=2mm,
    inner sep=1pt,
    font=\scriptsize\sffamily
  },
}

\tikzset{
  agentpath/.style={
    ->,
    thick,
    draw=agent
  },
}

\tikzset{
  agenttarget/.style={
    star,
    star points=5,
    star point ratio=2.25,
    draw=orange!80,
    thick,
    fill=orange!60,
    minimum size=3mm,
    inner sep=0pt
  },
}

\tikzset{
    bicycle/.style={
    draw=palette1!50!black, 
    fill=palette1, 
    minimum width=3mm, 
    minimum height=1.5mm, 
    transform shape
    }
}

%% file: 02_figures/fig01_overview.tex
\begin{figure*}
\centering
\begin{tikzpicture}[node distance=3mm and 3mm]

    \node[heading, anchor=west, minimum width=0.33\linewidth] (processesHead) at (6mm,0)
    {a) Processes};

    \node[minimum height=5mm, anchor= north west, minimum width = 45mm](p1Head) at (processesHead.south west)
    {Process $\pi_1$};

        \node[anchor=north, inner sep=0.5pt, anchor=north west] (poIPicture1) at ([xshift=7.2mm]p1Head.south west) {
            \begin{tikzpicture}
                \node[poi] (poi) {};
                \node[below=of poi] (n3) {};

                \draw[relation-undirected, dotted] (poi) -- (n3);
                
            \end{tikzpicture}
        };

        \node[flowchute, below=of poIPicture1, align=center] (queue) {$q_{\pi_1}$};

            \node[expl, right= of queue] (explArr) {
                    1) Objects arrive\\
                    at PoI.
            };

        \node[object, below left=1mm and 1mm of queue.south] (o1) {};
        \node[object, below=5mm of queue.south] (o2) {};

        \draw[<->, arrowlabel] ([xshift=3mm]o1.north) -- ([xshift=1mm]o2.north) node[midway,right]
        {$\Delta t$};

        \node[flowbox, below=of o2] (drain) {$d_{\pi_1}$};
            \node[expl, anchor=center] (explDrain) at (explArr|-drain) {
                    2) Objects drain\\
                    into DSG.
            };
    
        \draw[->, thick] (poIPicture1) -- (queue);
        
        \foreach \x in {o2}
          \draw[->, thick] (\x) -- (drain.north);

        \node[below=of drain, inner sep=0.5pt] (drainedPicture) {
            \begin{tikzpicture}[node distance=2mm and 2mm]
            
            \node[poi] (poi) {};
            
            \node[place, below =of poi] (p1) {};
            \node[place, right=of p1] (p2) {};
            \node[place, left =of p1] (p3) {};
            \node[place, above=of p2] (p4) {};
            \node[left=of p3] (p5){};
            \node[right=of p2] (p6){};
            
            \node[object, below left= of p2] (obj) {};
            
            \draw[relation-directed] (poi) -- (p1);
            \draw[relation-undirected] (p1) -- (p2);
            \draw[relation-undirected] (p1) -- (p3);
            \draw[relation-undirected] (p2) -- (p4);
            \draw[relation-undirected, dashed] (p3) -- (p5);
            \draw[relation-undirected, dashed] (p2) -- (p6);

            \draw[relation-directed] (p2) -- (obj);
            
            \draw[->, red] (poi) to[out=-120,in=120] (p1);
            \draw[->, red] (p1) to[out=30,in=170] (p2);
            \draw[->, red] (p2) to[out=-90,in=0] (obj);
            \end{tikzpicture}
        };

        \draw[->, thick] (drain.south) -- (drainedPicture.north);

        \node[flowbox, below= of drainedPicture] (lifecycle) {Lifecycle};
        \draw[->, thick] (drainedPicture.south) -- (lifecycle.north);
            \node[expl, right= of lifecycle] (explLife) {
                    3) Objects dwell\\
                    in DSG and are\\
                    deleted.
            };

        \node[below=of lifecycle, inner sep=0.5pt] (deletedPicture) {
            \begin{tikzpicture}[node distance=2mm and 2mm]
            
            \node[poi] (poi) {};
            
            \node[place, below =of poi] (p1) {};
            \node[place, right=of p1] (p2) {};
            \node[place, left =of p1] (p3) {};
            \node[place, above=of p2] (p4) {};
            \node[left=of p3] (p5){};
            \node[right=of p2] (p6){};
            
            \node[object, dashed, fill=none, below left= of p2] (obj) {};
            
            \draw[relation-directed] (poi) -- (p1);
            \draw[relation-undirected] (p1) -- (p2);
            \draw[relation-undirected] (p1) -- (p3);
            \draw[relation-undirected] (p2) -- (p4);
            \draw[relation-undirected] (p3) -- (p5);
            \draw[relation-undirected] (p2) -- (p6);

            \draw[relation-directed, dotted] (p2) -- (obj);
            
            \end{tikzpicture}
        };

        \draw[->, thick] (lifecycle.south) -- (deletedPicture.north);

        \node[right=of poIPicture1.north east, anchor=north west, inner sep=0, minimum height=0, minimum width=0] (otherPoIs) 
        {$\forall$ PoIs \tikz{\node[poi]{};} of class $\in \mathcal{C}_{q,\pi_1}$};

        \begin{pgfonlayer}{background}
            \node[process, fit=(poIPicture1)(drain)(lifecycle)(deletedPicture)(explArr)(explDrain)(explLife)(otherPoIs), inner sep=2pt] (p1) {};
        \end{pgfonlayer}

    \node[minimum height=5mm, anchor= north east](pnHead) at (processesHead.south east)
    {$\pi_N$};

        \node[process, minimum width=0, spanxy={pnHead}{p1}](pn) {};

    \node (ellipsis1) at ($ (p1Head.east)!0.5!(pnHead.west) $)  {$\cdots$};
    \node at (ellipsis1|-queue.center) {$\cdots$};
    \node at (ellipsis1|-drain.center) {$\cdots$};
    \node at (ellipsis1|-lifecycle.center) {$\cdots$};

    \begin{pgfonlayer}{backgroundGlob}
    \node[process, fit=(processesHead)(pn)(p1)] (processes) {};
    \end{pgfonlayer}


    \node[heading,anchor=east, minimum width=0.41\linewidth](agentsHead) at ([xshift=-25mm]\linewidth,0)
    {b) Multi-Agent Activity};

    \node[minimum height=5mm, anchor= north east, minimum width = 60.5mm](amHead) at (agentsHead.south east) {Agent $a_M$};

        \node[flowbox, anchor= north east] (fulfillTask) at ([xshift=-11mm, yshift=-1mm] amHead.south east) {Fulfill Task};

        \draw[<-, thick] (fulfillTask.east) -- ++(14mm,0) node[font=\scriptsize\sffamily, anchor=west](exTask){e.g., "Find \tikz[scale=.2]{\node[agenttarget, minimum size=1.5mm]{};}"};

        \node[flowbox, below=6mm of fulfillTask.south] (planning) {Navigate};
        \node[flowbox, anchor=east, minimum width=15mm] (extPlanning) at (exTask.east|-planning.center) {Decision \\ Making};
        \node[flowbox, below= 10mm of extPlanning , minimum width=15mm] (beliefModel) {Belief Model};
        \draw[->, thick] (extPlanning) -- (planning);
        \draw[->, thick] (extPlanning) -- (fulfillTask.south east);
        \draw[->, thick] (beliefModel) -- (extPlanning);

        \node[below=of planning, inner sep=0] (pathImage) {
            \begin{tikzpicture}
                \node[place] (p1) {};
                \node[place, right=of p1] (p2) {};
                \node[place, right=of p2] (p3) {};
                \node[place, right=of p3] (p4) {};
                \node[agenttarget, above right=of p4] (p5) {};
                \node[ below right=2mm of p4] (p6) {};
                
                \draw[relation-undirected] (p1) -- (p2) -- (p3) -- (p4);
                \draw[relation-undirected, dotted] (p4) -- (p5);
                \draw[relation-undirected, dotted] (p4) -- (p6);
                
                \node[poi, above=of p2] (poiA) {};
                \node[poi, above=of p4] (poiB) {};
                
                \draw[relation-directed] (p2) -- (poiA);
                \draw[relation-directed] (p4) -- (poiB);
                
                \node[object, below right=of p1] (o1) {};
                \node[object, below=of p3] (o2) {};
                
                \node[object, below left=of p1] (o3) {};
                \node[object, above left=of p1] (o4) {};
                \node[object, left=of o3] (o5) {};

                \draw[relation-directed] (p1) -- (o1);
                \draw[relation-directed] (p3) -- (o2);
                \draw[relation-directed] (p1) -- (o3);
                \draw[relation-directed] (p1) -- (o4);
                \draw[relation-directed] (o5) -- (o3);
                \draw[relation-directed] (o5) -- (o4);
                \draw[relation-directed] (p1) -- (o5);

                \node[mobileagent, anchor=center] (agent) at (p2){};
                \draw[agentpath] (agent) -- (p3) -- (p4) -- (p5);

            \end{tikzpicture}
        
        };

        \node[flowbox, below= of pathImage] (obsModel) {Obervation $\sigma_M$};
        \draw[<-, thick, dashed] (obsModel.east) -- ++(6mm,0) node[above, midway] {$\mathcal{G}[t]$};

        \node[below=of obsModel, inner sep=0] (obsImage){
            \begin{tikzpicture}
                \node[place] (p1) {};
                \node[place, right=of p1] (p2) {};
                \node[place, right=of p2] (p3) {};
                \node[place, right=of p3] (p4) {};
                \node[agenttarget, above right=of p4] (p5) {};
                \node[ below right=2mm of p4] (p6) {};
                
                \draw[relation-undirected] (p1) -- (p2) -- (p3) -- (p4);
                \draw[relation-undirected, dotted] (p4) -- (p5);
                \draw[relation-undirected, dotted] (p4) -- (p6);
                
                \node[poi, above=of p2] (poiA) {};
                \node[poi, above=of p4] (poiB) {};
                
                \draw[relation-directed] (p2) -- (poiA);
                \draw[relation-directed] (p4) -- (poiB);
                
                \node[object, below right=of p1] (o1) {};
                \node[object, below=of p3] (o2) {};
                
                \node[object, below left=of p1] (o3) {};
                \node[object, above left=of p1] (o4) {};
                \node[object, left=of o3] (o5) {};

                \draw[relation-directed] (p1) -- (o1);
                \draw[relation-directed] (p3) -- (o2);
                \draw[relation-directed] (p1) -- (o3);
                \draw[relation-directed] (p1) -- (o4);
                \draw[relation-directed] (o5) -- (o3);
                \draw[relation-directed] (o5) -- (o4);
                \draw[relation-directed] (p1) -- (o5);

                \node[circle, minimum size=8mm, anchor=center] (agentFOV) at (p2) {}; 
                \node[fit=(p1)(p2)(p3)(p4)(p5)(o1)(o2)(o3)(o4)(o5)] (fogArea) {};
                \fogoverlay{fogArea}{{agentFOV}}{4mm}{9mm}{agentsBackground!40}{0.85}

                \node[mobileagent, anchor=center] (agent) at (p2){};
                \draw[agentpath, -] (agent) -- (p3);
                \draw[agentpath, dashed] (p3) -- (p4) -- (p5);

            \end{tikzpicture}
        };

            \node[expl, left= 6mm of fulfillTask.north west, anchor=north east, minimum width=29mm] (explTask) {
                    1) Agents (\tikz[inner sep=0]{\node[mobileagent]{};})\\
                    fulfill tasks.
            };
            
            \node[expl, anchor=north, minimum width=29mm] (explPlanning) at (explTask.center|-planning.north) {
                    2) Task requires agents\\ to traverse paths (\tikz{\node[minimum height=3pt, inner sep=0, minimum width=0]{};\draw[agentpath, -] (0,0) -- ++(2mm,0);}).
            };

            \node[expl, anchor=north, minimum width=29mm] (explObs) at (explTask.center|-obsModel.north) {
                    3) During execution, \\
                    agents partially ob-  \\
                    serve the environment.
            };

        \draw[->, thick] (fulfillTask.south) -- (planning.north);
        \draw[->, thick] (planning.south) -- (pathImage.north);
        \draw[->, thick] (pathImage.south) -- (obsModel.north);
        \draw[->, thick] (obsModel.south) -- (obsImage.north);

        \begin{pgfonlayer}{background}
            \node[agent, fit=(fulfillTask)(planning)(pathImage)(obsModel)(obsImage)(explTask)(explPlanning)(explObs)] (am) {};
        \end{pgfonlayer}

    \node[minimum height=5mm, anchor= north west](a1Head) at (agentsHead.south west) {$a_1$};
    \node[agent, spanxy={a1Head}{am}, anchor=north west, minimum width=0] (a1) at (a1Head.south west) {};

    \node (ellipsis1) at ($ (a1Head.east)!0.5!(amHead.west) $) {$\cdots$};
    
    \node at (ellipsis1|-fulfillTask.center) {$\cdots$};
    \node at (ellipsis1|-planning.center) {$\cdots$};
    \node at (ellipsis1|-obsModel.center) {$\cdots$};

    \begin{pgfonlayer}{backgroundGlob}
    \node[agent, fit=(agentsHead)(a1)(am)] (agents) {};
    \end{pgfonlayer}


    \node[sgtrue, minimum height=20mm, minimum width=0.2\linewidth, anchor=north west, inner sep=2pt] (truesg) at ([xshift=0, yshift=-2mm] processes.south west) {%
        \begin{tikzpicture}
        \node[heading, anchor=north, inner sep=1pt, minimum height=0] (head) {c) True Dynamic Scene Graph $\mathcal{G}[t]$};
        \node[below=1pt of head.south, minimum height=0]{};
        \end{tikzpicture}\\
        \begin{tikzpicture}[node distance=3mm, inner sep=0]
        \node[place] (p1) {};
        \node[place, right=of p1] (p2) {};
        \node[place, right=of p2] (p3) {};
        \node[place, right=of p3] (p4) {};
        \node[place, above right=of p4] (p5) {};
        \node[place, below right=of p4] (p6) {};
        
        \draw[relation-undirected] (p1) -- (p2) -- (p3) -- (p4);
        \draw[relation-undirected] (p4) -- (p5);
        \draw[relation-undirected] (p4) -- (p6);
        
        \node[poi, above=of p2] (poiA) {};
        \node[poi, above=of p4] (poiB) {};
        \node[poi, right=of p6] (poiC) {};
        
        \draw[relation-directed] (p2) -- (poiA);
        \draw[relation-directed] (p4) -- (poiB);
        \draw[relation-directed] (p6) -- (poiC);
        
        \node[object, below right=of p1] (o1) {};
        \node[object, below=of p3] (o2) {};
        
        \node[object, below left=of p1] (o3) {};
        \node[object, above left=of p1] (o4) {};
        \node[object, left=of o3] (o5) {};
        
        \node[object, right=of p5] (o6) {};
        \node[object, above right =of poiC] (o7) {};
        \node[object, right=of poiC] (o8) {};
        
        \draw[relation-directed] (p1) -- (o1);
        \draw[relation-directed] (p3) -- (o2);
        \draw[relation-directed] (p1) -- (o3);
        \draw[relation-directed] (p1) -- (o4);
        \draw[relation-directed] (o5) -- (o3);
        \draw[relation-directed] (o5) -- (o4);
        \draw[relation-directed] (p1) -- (o5);
        \draw[relation-directed] (p5) -- (o6);
        \draw[relation-directed] (poiC) -- (o7);
        \draw[relation-directed] (poiC) -- (o8);
        \draw[relation-directed] (o7) to[out=-100,in=130] (o8);
        \draw[relation-directed] (o8) to[out=80,in=-60] (o7);

        \node[anchor=north west, minimum width=0, minimum height=0, inner sep=0, xshift=4mm] (legendHead) at (o8.east |- poiB.north) {Node/ Edge Types:};

            \node[place, below=2mm of legendHead.south west, anchor=west] (p7) {};
            \node[place, right=2mm of p7] (p8) {};
            \draw[relation-undirected] (p7) -- (p8);
            \node[anchor=west, right=2mm of p8.east, inner sep=0, minimum width=0, minimum height=0] (pathNetworkEx) {Path Network $\mathcal{P}$};

            \node[object, below=1mm of p7] (o9){};
            \node[anchor=west, inner sep=0, minimum width=0, minimum height=0, anchor=west] (objectEx) at (pathNetworkEx.west|-o9.center) {Objects};

            \node[poi, below=1mm of o9] (poiD){};
            \node[anchor=west, inner sep=0, minimum width=0, minimum height=0, anchor=west] (poiEx) at (pathNetworkEx.west|-poiD.center) {Point of Interest};

        \end{tikzpicture}
    
    };

    \node[flowbox, align=center, anchor=north west] (sginit) at ([xshift=24mm, yshift = -2mm]truesg.south west) 
    {SG Initialization};

        \node[anchor=west] (osm) at ([xshift=-24mm]sginit.west) {};
        \draw[->, dashed, thick] (osm) -- node[below]{e.g., OSM data} (sginit.west);

        \draw[->, thick] (sginit.east) -| node[below]{$\mathcal{G}_0$} ([xshift=10mm]truesg.south);

    \coordinate (truesgBendL) at ([xshift=-8mm,yshift=0mm]truesg.west);
    \coordinate (truesgBendS) at ([xshift=-3mm,yshift=0mm]truesg.west);
    \coordinate (innerSGWest) at ([xshift=0mm,yshift=5mm]truesg.west);
    
        \draw[->, thick](truesg.west) -| (truesgBendL)  |- ([yshift=1.5mm]poIPicture1.west) node[xshift=3.5mm, midway, above, sloped, ann] {$\mathcal{G}[t]$} ;
        \node[ann, yshift=-1mm, anchor=north west]  at (truesgBendL|-truesg.west) {uses};
        
        \draw[->, thick](truesg.west) -| (truesgBendL) |- (drain.west) node[xshift=3.5mm, midway, above, sloped, ann] {$\mathcal{G}[t]$} ;
        
        \draw[->, thick](drainedPicture.west) -| ([yshift=10mm]truesgBendS) |- (innerSGWest) ;
        
        \draw[->, thick](deletedPicture.west) -| node[anchor=south, yshift=2mm, midway, above, ann]{updates} ([yshift=10mm]truesgBendS) |- (innerSGWest) ;


    \node[sgobs, minimum height=1.5cm, minimum width=0.41\linewidth, anchor=north west] (obssg) at (agents.south west|-truesg.north east) {
    \begin{tikzpicture}
        \node[heading, anchor=north, inner sep=0, minimum height=0] {d) Observed Dynamic Scene Graph $\mathcal{G}_\text{obs}[t]$};
    \end{tikzpicture}\\
    \begin{tikzpicture}[node distance=3mm, inner sep=0]
        \node[place] (p1) {};
        \node[place, right=of p1] (p2) {};
        \node[place, right=of p2] (p3) {};
        \node[place, right=of p3] (p4) {};
        \node[place, above right=of p4] (p5) {};
        \node[place, below right=of p4] (p6) {};
        
        \draw[relation-undirected] (p1) -- (p2) -- (p3) -- (p4);
        \draw[relation-undirected] (p4) -- (p5);
        \draw[relation-undirected] (p4) -- (p6);
        
        \node[poi, above=of p2] (poiA) {};
        \node[poi, above=of p4] (poiB) {};
        \node[poi, right=of p6] (poiC) {};
        
        \draw[relation-directed] (p2) -- (poiA);
        \draw[relation-directed] (p4) -- (poiB);
        \draw[relation-directed] (p6) -- (poiC);
        
        \node[object, below right=of p1] (o1) {};
        \node[object, below=of p3] (o2) {};
        
        \node[object, below left=of p1] (o3) {};
        \node[object, above left=of p1] (o4) {};
        \node[object, left=of o3] (o5) {};
        
        \node[object, right=of p5] (o6) {};
        \node[object, above right =of poiC] (o7) {};
        \node[object, right=of poiC] (o8) {};
        
        \draw[relation-directed] (p1) -- (o1);
        \draw[relation-directed] (p3) -- (o2);
        \draw[relation-directed] (p1) -- (o3);
        \draw[relation-directed] (p1) -- (o4);
        \draw[relation-directed] (o5) -- (o3);
        \draw[relation-directed] (o5) -- (o4);
        \draw[relation-directed] (p1) -- (o5);
        \draw[relation-directed] (p5) -- (o6);
        \draw[relation-directed] (poiC) -- (o7);
        \draw[relation-directed] (poiC) -- (o8);
        \draw[relation-directed] (o7) to[out=-100,in=130] (o8);
        \draw[relation-directed] (o8) to[out=80,in=-60] (o7);

        \node[circle, minimum size=8mm, anchor=center] (agentFOV) at (p2) {}; 
        \node[minimum width=0, fit=(p1)(p2)(p3)(p4)(p5)(o1)(o2)(o3)(o4)(o5)(o6)(o7)(o8)] (fogArea) {};
        \fogoverlay{fogArea}{agentFOV, p6}{4mm}{6mm}{obsSGBackground!50}{0.85}

        \node[mobileagent, anchor=center] (agent) at (p2){};
        \node[mobileagent, anchor=center] (agent2) at (p6){};    

        \node[expl, minimum width=0, minimum height =0, anchor=east] at ([xshift=-2mm]fogArea.west){
            \parbox{25mm}{Partial Observability inherently introduces differences between $\mathcal{G}[t]$ and $\mathcal{G}_\text{obs}[t]$.}
        };
    \end{tikzpicture}
    };

        \draw[->, thick] (obsImage.south|-am.south) -- (obsImage|-obssg.north) node[xshift=1mm, yshift=-1mm, ann, midway, right, ann]{updates};
        \draw[->, thick] (a1.south) -- (a1.south|-obssg.north) node[xshift=1mm, yshift=-1mm, ann, midway, right]{updates};

        \draw[->, thick] (obssg.east) -| (beliefModel);
        \node[ann, anchor=north east, yshift=-1mm ] at (beliefModel|-obssg.east) {uses};
        \node[ann, anchor=north west, xshift=2pt] at (beliefModel.south) {$\mathcal{G}_\text{obs}$};
    
        \draw[->, thick] (sginit.east) -| node[right]{$\mathcal{G}_{\text{obs},0} :=\mathcal{G}_0$} (obssg.south);

\end{tikzpicture}
\caption{Overview of \fogmachine: The environment is represented as a DSG (c), updated by arrival and deletion events from processes (a). Mobile agents (b) perform tasks in this \ac{dsg}-structured environment and, in doing so, observe parts of it, yielding the observed DSG (d).}

\label{fig:overview}
\end{figure*}
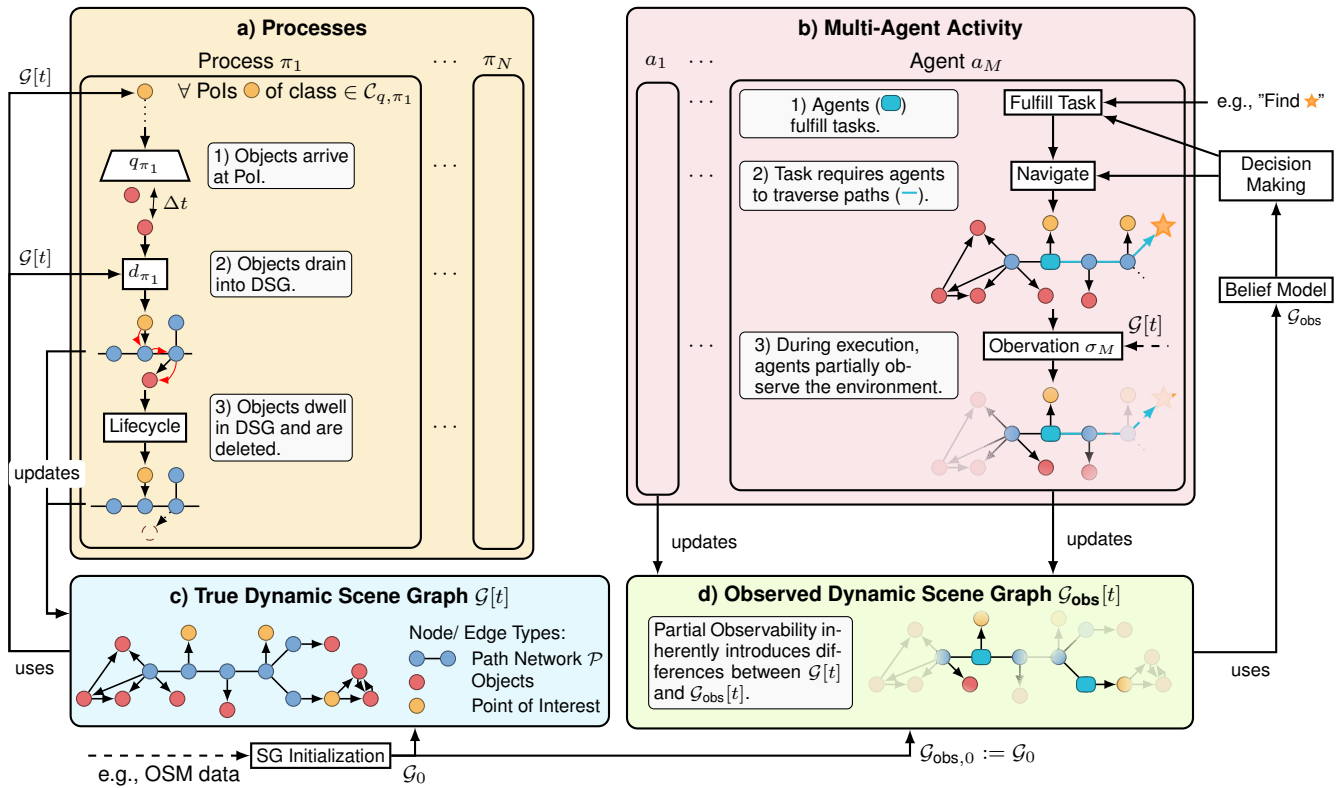

%% file: 01_sections/00_abstract.tex
Dynamic Scene Graphs (DSGs) provide a structured representation of hierarchical, interconnected environments, but current approaches struggle to capture stochastic dynamics, partial observability, and multi-agent activity. These aspects are critical for embodied AI, where agents must act under uncertainty and delayed perception.
We introduce \fogmachine, an open-source framework that fuses DSGs with discrete-event simulation to model object dynamics, agent observations, and interactions at scale. This setup enables the study of uncertainty propagation, planning under limited perception, and emergent multi-agent behavior. Experiments in urban scenarios illustrate realistic temporal and spatial patterns while revealing the challenges of belief estimation under sparse observations. By combining structured representations with efficient simulation, \fogmachine establishes an effective tool for benchmarking, model training, and advancing embodied AI in complex, uncertain environments.

%% file: 01_sections/01_introduction.tex
\section{Introduction}\label{sec:introduction}

Despite advances in embodied AI, autonomous agents still struggle in environments that are only partially observable and evolve unpredictably. To act robustly, agents must connect local perception with global reasoning, yet most approaches treat these levels in isolation.  

\acp{sg} have emerged as a powerful abstraction, encoding objects, infrastructure, and places into relational structures for embodied agents~\cite{rosinol3DDynamicScene2020, saucedoBeliefSceneGraphs2024}. They support reasoning beyond raw sensor data, but are typically used as static world models. Generative approaches in this space lack mechanisms for capturing \emph{plausible, stochastic dynamics} that characterize real environments. Moreover, embodied agents can never observe the entire world at once---changes in the \ac{sg} only become actionable once perceived. This \emph{foggy} reality underscores the need to explicitly model both \emph{partial observability} and \emph{delayed perception} in agent--environment interaction.

\ac{des} provides a complementary perspective: it efficiently models complex, asynchronous processes inside large systems in domains such as logistics and manufacturing. \ac{des} is well suited to capture event-driven dynamics, still its integration with robotics and embodied AI has remained limited. This highlights an opportunity: the discrete evolution of \acp{sg} topology can naturally be modeled within a \ac{des}, while partial observability can be represented through uncertainty-aware masking of the true state.  

We present \fogmachine\footnote{\textbf{F}used \textbf{O}bserved Scene \textbf{G}raphs from \textbf{M}ulti‑Agent \textbf{Ac}tivity in \textbf{H}ierarchical, \textbf{In}terconnected \textbf{E}nvironments}, a simulation framework that fuses \ac{des} with \ac{sg} representations. Environments are modeled as \acp{dsg} whose nodes and edges evolve through stochastic or deterministic processes. Mobile agents traverse task-specific paths and acquire local observations, enabling the study of \emph{uncertainty propagation} and \emph{planning loops} in hierarchical, interconnected environments. We demonstrate \fogmachine in the domain of \acp{adr}, where no \ac{dsg}-based datasets exist. Using \ac{osm} data~\cite{OpenStreetMap}, we \emph{automatically generate \ac{dsg} representations of real-world urban areas} and conduct extensive simulation studies across three \acp{sg} drawn from German cities.  

\textbf{To the best of our knowledge, \fogmachine is the first open-source\footnote{repository available soon} framework that applies discrete-event simulation to dynamic scene graphs, providing an efficient tool for studying embodied AI under partial observability.}

This work is structured as follows: First, we report relevant related work in \cref{sec:related_work}. Then we introduce the core modeling approaches and components of \fogmachine in \cref{sec:methods}. Using the resulting framework, we provide an application example in the domain of \acp{adr} in \cref{sec:application}. We discuss the results in \cref{sec:conclusion} and conclude with an outlook on possible future directions in this field. 

%% file: 01_sections/02_related_work.tex
\section{Related Work}\label{sec:related_work}
\acp{dsg} are a well discussed topic in scientific literature. For the purpose of this work, we report research that relates to \fogmachine and \acp{dsg} in the following ways: (A) literature which emphasizes the need for \ac{dsg}-generative models and (B) existing research on \ac{dsg}-generative models.

\subsection{Need for \ac{dsg}-Generative Models}  
Recent works have highlighted the necessity of generative models for \acp{dsg} that go beyond static perception and incorporate realistic system dynamics. \textcite{heuerBenchmarkingMultiRobotCoordination2024} point out the lack of structured benchmarks for graph-based reasoning in robotics, suggesting that simulation could provide controlled evaluation environments. 
We see different upstream and downstream tasks which could benefit from a realistic \ac{dsg}-generative model. Scene generation can be conditioned using \acp{sg}. For example, \textcite{liuControllable3DOutdoor2025} use \acp{sg} of urban environments as inputs to generate 3d environments. Further conditioning these scene-generative models using temporally consistent \acp{sg} could improve realism. 
Downstream planning and decision making can also benefit from an efficient generative model. \textcite{kurenkovModelingDynamicEnvironments2023} study dynamic link prediction on evolving \acp{sg}, a task which emphasizes the need for modeling uncertain and partial world knowledge that can naturally be benchmarked on \fogmachine with realistic stochastic updates.  Similarly, \textcite{guConceptGraphsOpenVocabulary3D2024} propose concept graphs for testing downstream applications, which could be stress-tested with simulated \acp{dsg}. Navigation tasks can also use \acp{dsg} as inputs. Research employs \acp{sg} within \acp{pomdp}, where uncertainty and partial observability are central. Approaches such as SARP for context-aware planning under occlusion~\cite{amiriReasoningSceneGraphs2022}, 3DSG-POMDPs with Bayesian belief updates~\cite{remy2023semanticallydriven}, and graph-based RL navigation frameworks~\cite{oskolkovSGNCIRLSceneGraphbased2025, ravichandranHierarchicalRepresentationsExplicit2022} exemplify how structured world models can support decision-making under uncertainty. A simulator with inferable stochastic dynamics, e.g., via Monte Carlo sampling, would provide a valuable platform for training and benchmarking such methods.
Together, these works motivate the need for simulation environments that can produce dynamic, hierarchical graphs under uncertainty.  

\subsection{Existing Generative Models for \acp{dsg}}  
A number of efforts benchmark on scene graphs, but often fall short of modeling their stochastic evolution. Habitat~\cite{savvaHabitatPlatformEmbodied2019} integrates scene graphs to model semantic structure of indoor environments, yet scene graph dynamics are dataset-driven rather than simulated. Hou \emph{et al.}~\cite{houHiDynaGraphHierarchical2025} employ hierarchical scene graphs to capture both local and global object relations for downstream tasks, but again treat the graphs as inputs rather than as evolving states. The Dynamic House Simulator by Kurenkov \emph{et al.}~\cite{kurenkovModelingDynamicEnvironments2023} provides evolving scene graphs, though its dynamics are driven by simple stochastic update rules. Nayak \emph{et al.}~\cite{nayakLLaMARLongHorizonPlanning2025} present MAP-THOR, a benchmark dataset for multi-agent tasks in AI2-THOR, but the graph evolution remains implicit in agent interactions.  

Datasets for \ac{dsg} research have also emerged. 4D-OR~\cite{ozsoyHolisticDomainModeling2024} provides 3D semantic scene graphs for surgical scenarios, with dynamics captured through sensor data and manual annotation. STAR~\cite{liSTARFirstEverDataset2025} targets satellite imagery, offering static scene graphs without temporal evolution. Wang \emph{et al.}~\cite{wangIndVisSGGVLMbasedScene2025} annotate spatio-temporal scene graphs from image sequences, but lack generative capabilities.
While these efforts demonstrate demand for such world models, none provide a controlled simulation of evolving, uncertain environments.

%% file: 01_sections/03_methods.tex
\section{Methods}\label{sec:methods}
As discussed above, \acp{dsg} are natural representations of hierarchical environments. We leverage this powerful model to describe the environment and the agents interacting with it. The overall architecture of \fogmachine can be broken down into several key components: (1) The \ac{dsg} used to represent the environment - including a traversable path network for mobile agents to move in, (2) processes that manipulate the \ac{dsg} by spawning and deleting objects, and (3) mobile agents that fulfill tasks that require them to interact with the \ac{dsg} and during which they observe parts of the world. These components are then integrated into a discrete-event simulation. An overview of the architecture is depicted in \cref{fig:overview}.

\input{02_figures/fig02_experiment}

\subsection{Dynamic Scene Graph}
The core environmental representation of  \fogmachine  is a global scene graph $\mathcal{G}[t]= \left(\mathcal{V}[t], \mathcal{E}[t]\right)$ which varies at discrete times $t$. $\mathcal{G}[t]$ contains different types of nodes $v^{(i)}[t] \in \mathcal{V}[t]$ and edges $e^{(j)}[t] \in \mathcal{E}[t]$. Analogously to \cite{houHiDynaGraphHierarchical2025}, we differentiate between a static subgraph $\mathcal{S}\subseteq \mathcal{G}[t]$ and a dynamic subgraph $\mathcal{D}[t]$. We further subdivide the scene graph $\mathcal{G}[t]$ to model a traversable path network $\mathcal{P} \subseteq \mathcal{S}$ and objects as well as their relations.

\textbf{Traversable Path Network} Mobile agents can navigate the simple, (un)directed topometric network $\mathcal{P} = \left(\mathcal{V}_{P}, \mathcal{E}_P, w\right)$ to observe the environment and complete assigned tasks. Importantly, we do not consider the network's topology as dynamic. Instead, we encode possible restrictions, or severed connections via cost functions $w(\cdot)$ associated with edges and nodes, triggered by objects related to them. Long-term changes that alter the topology of the path network are considered to be of minor influence in comparison to object-driven alterations and are hence neglected. Nodes $v_P^{(i)} \in \mathcal{V}_P$ are connected using (un)directed adjacency relations $e_P^{(j)} \in \mathcal{E}_P$. Furthermore, we allow hierarchical relations to other nodes.

\textbf{Objects} We consider different classes of objects that are observed by and interact with mobile agents, their tasks and the environment. Each object $o^{(i)}=\left(c_j, t_\text{spawn}^{(i)}, t_L^{(i)}\right) \in \mathcal{V}_O \subset \mathcal{V}$ therefore is assigned a semantic class $c_j \in \mathcal{C}_V$. These objects can be any observable entity with a lifecycle. While the overall approach is applicable to different object dynamics, it is most powerful for semi-static objects (objects that enter the system and then remain stationary until exiting the system). The lifecycle is described by the arrival time $t_\text{spawn}^{(i)}$ and life time $t_L^{(i)}$ and controlled through the processes described below. The objects' lifecycle introduces dynamics to the otherwise static \ac{sg}. 
Objects are connected to the \ac{dsg} through a set of problem-specific edges. These could model hierarchical relations (e.g., \texttt{contains}) or neighborhood (e.g., \texttt{next\_to}). Importantly, objects are not part of the path network ($\mathcal{V}_O \cap \mathcal{V}_P = \varnothing$ ). This restriction allows for a clear separation between objects and infrastructure.

\subsection{Processes}
We manipulate the dynamic components of the \ac{dsg} through rule-based, event-discrete, stochastic or deterministic processes $\Pi$. A process $\pi \in \Pi$ is defined as the tuple
\begin{equation}
   \pi=\left(\mathcal{C}_{q,\pi}, \mathcal{C}_{d,\pi}, q_\pi, d_\pi, l_\pi \right). 
\end{equation}
$\pi$ aims to spawn objects $o_\pi$ of semantic class $c_d \in \mathcal{C}_{d,\pi} \subseteq \mathcal{C}_V$ at a so called \ac{poi} node. We consider a node $v_\text{PoI}$ to be of interest for process $\pi$ if the node has a semantic class $c_s \in \mathcal{C}_{q,\pi} \subseteq \mathcal{C}_V$. For example, both a kitchen and a dining room would be of interest for a plate-spawning process. The process $\pi$ then creates arrival events at the \ac{poi} with inter-arrival times $\Delta t$ sampled from a process- and \ac{poi}-specific deterministic or stochastic process defined by the source function $q_\pi$ with 
\begin{equation}
    \Delta t = q_\pi \left(t, \mathcal{G}[t], v_\text{PoI}\right).
\end{equation}
Objects spawned this way then traverse $\mathcal{G}[t]$ to attach to a node in the neighborhood of the \ac{poi} node. This object passing mimics real-world causal relations and provides a  realistic  emergent behavior. The passing is described by the drain function $d_\pi$ with
\begin{equation}
    \left(o_\pi, \mathcal{E}_\pi\right) = d_\pi \left(t, \mathcal{G}[t], v_\text{PoI} \right).
\end{equation}
Hence, we receive a new object $o_\pi$ that is added to the \ac{dsg} and connected via a set $\mathcal{E}_\pi$ edges to existing nodes. In the simplest case, the object directly attaches to the \ac{poi}.

As most of the objects have a finite dwell time within the \ac{dsg}, we need to account for deletion. This is modeled by the lifetime attribute $t_L$ that again is sampled from a process- and object-specific deterministic or stochastic process with a lifetime function 
\begin{equation}
    t_L = l_\pi \left(t, \mathcal{G}[t], o_\pi \right).
\end{equation}
When objects exceed their lifetime, they are removed from the \ac{dsg}. 

The co-design of source, drain and lifetime functions enables the modeling of causal relations. For example, moving objects can be drained from one node and directly appear on an adjacent node. Also, for objects that are mostly hidden from the interactable environment (e.g.\,, cars parked inside closed garages), we can employ an open system strategy to improve model efficiency. 

\subsection{Mobile Agents, Tasks and Observations}
\label{subsec:mobile_agents_tasks_observations}
Next, we'll integrate a fleet $\mathcal{A}$ of mobile agents $a_i\in \mathcal{A}$ into the structure, resulting in the observed \ac{dsg}, $\mathcal{G}_\text{obs}[t]$.

An agent fulfills specific tasks. While these may be different on a semantic level, they will most certainly involve movement through the environment. We model this movement along a valid task-specific path $p=\left<v_1,\ldots,v_n\right>, \, v_i \in \mathcal{V}_P$. This path can be dynamically recomputed based on information gained from the world model. Due to the aforementioned cost functions, traversing nodes and edges takes time. At every node traversal, two discrete observations, $\sigma_i\left[t_{in}\right]$, $\sigma_i\left[t_{\text{out}}\right]$ are performed, where the agent perceives a (noisy) part of the true state of the \ac{dsg}. The observation is governed by the observability function $\sigma_i$ of the agent $a_i$. $\sigma_i$ encodes limitations of the agent's perception system like sensor range and epistemic uncertainty inside the perception pipeline. In the simplest case, an observation is a subgraph of the \ac{dsg}:
\begin{equation}
    \sigma_i\left(\mathcal{G}[t] \right) \subseteq \mathcal{G}[t].
\end{equation}
The observations of each agent can then be used to update a shared world model of the agent fleet, the observed \ac{dsg}, $\mathcal{G}_\text{obs}[t]$. This update $\oplus$ commonly requires alignment and plausibility checks of the local observations with the global graph. For simplicity, we will omit this step here and discuss potential directions in \cref{sec:conclusion}. In summary, observations provide updates of the observed \ac{dsg}, $\mathcal{G}_\text{obs}[t]$:
\begin{equation}
    \mathcal{G}_\text{obs}[t_{i}] = \mathcal{G}_\text{obs}[t_{i-1}] \oplus \sigma_i\left(\mathcal{G}[t_i] \right). 
\end{equation}
An update may involve adding or removing nodes and edges (that are part of the dynamic subgraph), as well as changing node and edge attributes.

\subsection{Discrete-Event Simulation}
\fogmachine employs \ac{des} as a temporal generative model to evolve the \ac{sg} under event-driven deterministic and stochastic processes. The \ac{des} is seeded using an external, static \ac{sg}, $\mathcal{G}_0$ without objects, $\mathcal{D}=\varnothing$. For simplicity, the observed \ac{sg} is seeded equally, $\mathcal{G}_\text{obs}[t_0] := \mathcal{G}_0$.
Then, each of the components above introduces events:

\textbf{Processes} attach events to \acp{poi} to spawn new objects. Spawn events are spaced by the inter-arrival times drawn from the source function $q_\pi$. After the objects attach to a node (see drain function $d_\pi$), the life cycle of the object triggers an event after reaching its specified life-time $t_L$.

\textbf{Mobile Agents} trigger observation events when entering and exiting nodes of the traversable path network $\mathcal{P}$. These events are spaced using the required travel times through nodes and edges (see cost function $w$).

\textbf{Tasks} depend on the concrete use case. However, we can assume without loss of generality that tasks are considered processes with deterministic or stochastic inter-arrival times. Tasks are assigned to agents which then trigger movement.

\subsection{Stochastic Models}\label{sec:stoch_models}
A \ac{nhpp} with  time-variant rate $\lambda(t)$ is a commonly used model for inter-arrival times. In \cref{sec:application}, we employ \acp{nhpp} for various components. To efficiently compute inter-arrival times for \acp{nhpp}, we employ a \textbf{Thinning} approach \cite{lewisSimulationNonhomogeneousPoisson1979}.
The global scene graph is conceptualized as an open system, embedded within a larger closed system. While the number of objects in the scene graph may vary over time, the total number of objects in the closed system remains constant. This closed system assumption becomes increasingly valid as the size of the system grows, as relatively fewer objects leave or enter the system. Within the open system, the lifetime of an individual object is governed by its type-specific mean lifetime, as well as by the arrival distributions of objects of the same type at other \acp{poi} within the closed system ensuring a balance of objects alive. 

\subsection{Monte Carlo Sampling} Since the underlying processes—arrivals, movements, and observations—are defined only implicitly through event rules, closed-form distributions of graph states are generally intractable. By repeatedly executing the \ac{des}, we can obtain empirical samples of scene graph trajectories. These samples can be aggregated to approximate complex distributions of interest, such as the probability of encountering an object of a given type at nodes of a particular class, or the expected temporal persistence of structural relations. In this way, the \ac{des} acts as a Monte Carlo estimator for properties of the dynamic, partially observed graph. These could then be used as improved inputs for methods such as \cite{kurenkovModelingDynamicEnvironments2023}.

%% file: 02_figures/fig02_experiment.tex
\begin{figure*}
\centering
\begin{tikzpicture}[
    every node/.style={inner sep=0,outer sep=0}
]

\node[minimum width=6cm, minimum height=5cm, anchor=south west] (heatmap) at (3mm,0)
    {\includegraphics[width=0.47\linewidth, trim=140mm 40mm 42mm 60mm, clip]{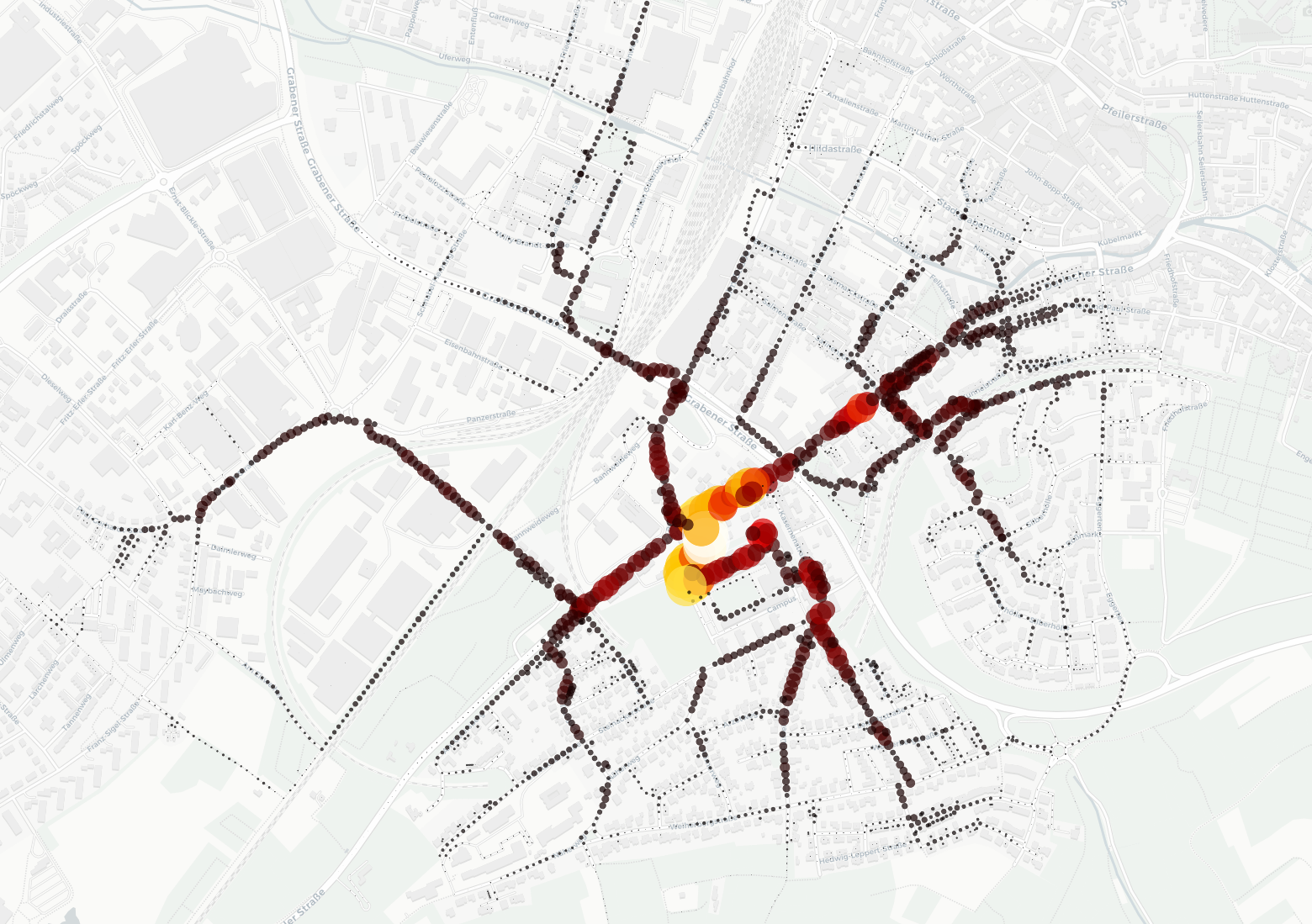}};
\node[above=2mm of heatmap, anchor=south] {b) Observation Heatmap Bruchsal (Boundaries clipped)};

\node[] (depotNode) at ([xshift=-6.5mm, yshift=-5mm] heatmap.center) {};
\node[inner sep=2pt] (depotMarker) at ([xshift=-10mm, yshift=1mm] depotNode.center) {\footnotesize Depot};
\draw[] (depotNode) -- (depotMarker);

\node[] (node1) at ([xshift=9.5mm, yshift=22mm] heatmap.center) {};
\node[] (marker1) at ([xshift=-7mm, yshift=2mm] node1.center) {\circnum{1}};
\draw[] (node1) -- (marker1);

\node[] (node2) at ([xshift=-5mm, yshift=-8mm] heatmap.center) {};
\node[] (marker2) at ([xshift=-3mm, yshift=-6mm] node2.center) {\circnum{2}};
\draw[] (node2) -- (marker2);

\node[] (node3) at ([xshift=37mm, yshift=10.6mm] heatmap.center) {};
\node[] (marker3) at ([xshift=6mm, yshift=1mm] node3.center) {\circnum{3}};
\draw[] (node3) -- (marker3);

\node[ minimum width=0.47\linewidth, minimum height=65mm, 
      anchor=north east] (subgraph) at ([xshift=-3mm]heatmap.north west){};
\node[above=0.1cm of subgraph] {a) Local Subgraph at \circnum{1}};

    \begin{scope}
        \clip (subgraph.south west) rectangle (subgraph.north east);
        \node at (subgraph.center) {\includegraphics[width=0.49\linewidth]{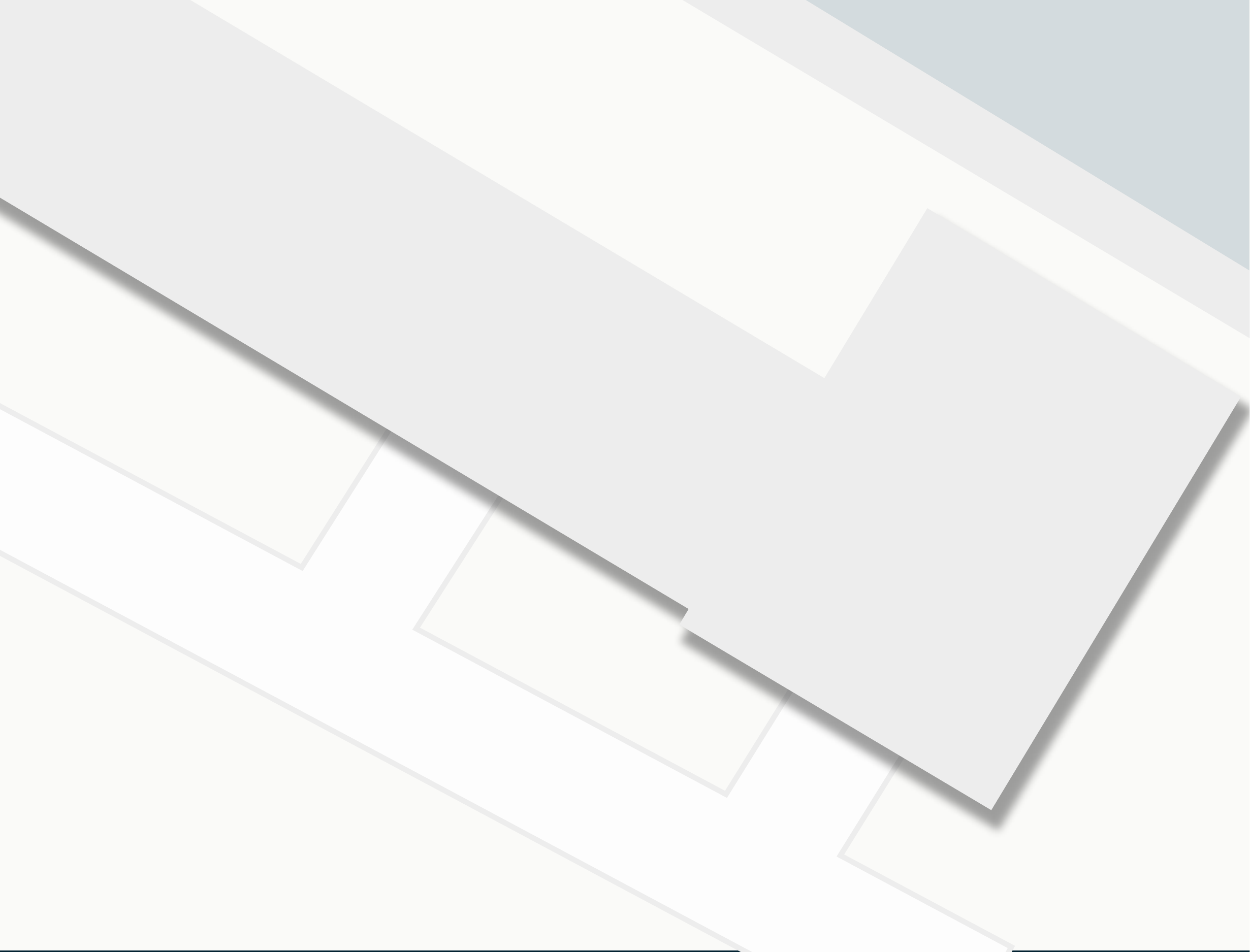}};

    \node[poi] (schoolPoI) at ([xshift=-12mm, yshift=12mm] subgraph.center){};
    \node[above= 0mm of schoolPoI] {Highschool};

    \node[place] (p1) at ([xshift=-3mm, yshift=-14mm] schoolPoI) {};
    \node[place] (p2) at ([xshift=-6mm, yshift=-10mm] p1) {};
    \node[place] (p3) at ([xshift=-20mm, yshift=11mm] p2) {};
    \node[place] (p4) at ([xshift=20mm, yshift=-11mm] p2) {};
    \node[place] (p5) at ([xshift=20mm, yshift=-11mm] p4) {};

    \node[left=1mm of p1] {\circnum{1}};

    \begin{scope}[shift={(p1.center)}, rotate=-32]
        \node[bicycle] (b1) at (5mm, 2mm) {};
        \node[bicycle] (b2) at (5mm, -1mm) {};
        \node[bicycle] (b3) at (5mm, -4mm) {};
        
    \end{scope}
    \node[right= 1mm of b3] {Bicycles};

    \node[object] (o1) at (b1){};
    \node[object] (o2) at (b2){};
    \node[object] (o3) at (b3){};

    \draw[relation-directed] (p1) -- (schoolPoI) node[right, midway, font=\scriptsize\sffamily, xshift=1mm] {\texttt{connects\_to}};

    \draw[relation-undirected] (p1) -- (p2);
    \draw[relation-undirected] (p3) -- (p2) -- (p4) -- (p5);

    \draw[relation-directed] (o1) -- (p1);
    \draw[relation-directed] (o2) -- (p1);
    \draw[relation-directed] (o3) -- (p1) node[below, midway,font=\scriptsize\sffamily, xshift=-4pt]{\texttt{on}};

    \end{scope}

\node[anchor=north east, fill=gray!10, inner sep=2pt] (subgraphLegend) at ([yshift=-1mm, xshift=-1mm] subgraph.north east) {
\begin{tikzpicture}[baseline=(current bounding box.north)]
        \node[anchor=north west, minimum width=0, minimum height=0, inner sep=0] (legendHead) at (o8.east |- poiB.north) {Node/ Edge Types:};

            \node[place, below=2mm of legendHead.south west, anchor=west] (p7) {};
            \node[place, right=2mm of p7] (p8) {};
            \draw[relation-undirected] (p7) -- (p8);
            \node[anchor=west, right=2mm of p8.east, inner sep=0, minimum width=0, minimum height=0] (pathNetworkEx) {Path Network $\mathcal{P}$};

            \node[object, below=1mm of p7] (o9){};
            \node[anchor=west, inner sep=0, minimum width=0, minimum height=0, anchor=west] (objectEx) at (pathNetworkEx.west|-o9.center) {Objects};

            \node[poi, below=1mm of o9] (poiD){};
            \node[anchor=west, inner sep=0, minimum width=0, minimum height=0, anchor=west] (poiEx) at (pathNetworkEx.west|-poiD.center) {Point of Interest};
            
\end{tikzpicture}
};


\node[anchor=north, minimum width=0.32\linewidth](plot2cap) at (0, -4mm)
    {d) Arrival Rate at \circnum{2}};
\node[anchor=north](plot2) at ([yshift=-2mm] plot2cap.south) 
    {\includegraphics[scale=1.0]{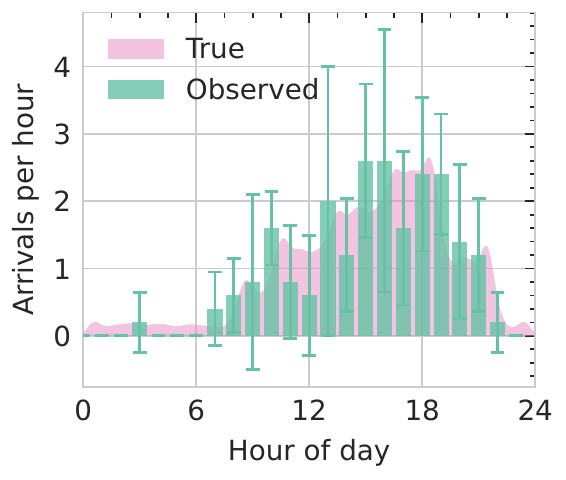}};
\node[emphbox, anchor=north, minimum width=0.32\linewidth](plot2interobs) at ([yshift=-0mm] plot2.south)
    {Interobs. Time of \circnum{2}: \numWithUncertaintyFromCSV[round-mode=places,round-precision=2]{04_media/observations_c562c745-7677-4ede-bba2-3ecc06d0eb01.csv}{interObservationTime}{\second}};
    
\node[anchor=north east, minimum width=0.32\linewidth](plot1cap) at ([xshift=-2mm] plot2cap.north west)
    {c) Arrival Rate at \circnum{1}};
\node[anchor=north](plot1) at  ([yshift=-2mm] plot1cap.south) 
    {\includegraphics[scale=1.0]{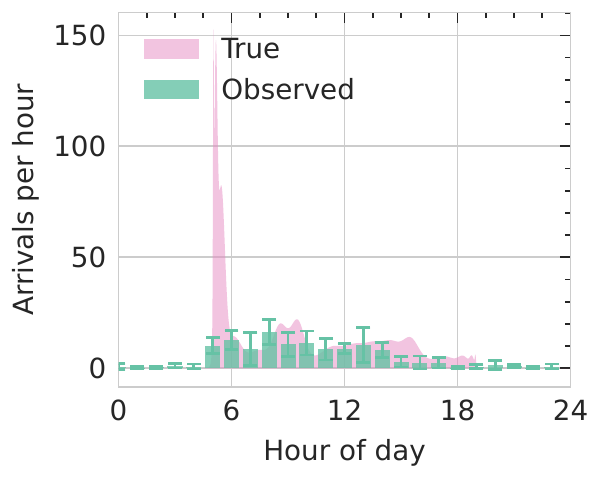}};
\node[emphbox, anchor=north, minimum width=0.32\linewidth](plot1interobs) at ([yshift=-0mm] plot1.south)
    {Interobs. Time of \circnum{1}: \numWithUncertaintyFromCSV[round-mode=places,round-precision=2]{04_media/observations_5d10ce87-81a9-4306-b889-95503f8b6ea4.csv}{interObservationTime}{\second}};
    
\node[anchor=north west, minimum width=0.32\linewidth](plot3cap) at ([xshift=2mm] plot2cap.north east) 
    {e) Arrival Rate at \circnum{3}};
\node[anchor=north](plot3) at ([yshift=-2mm] plot3cap.south) 
    {\includegraphics[scale=1.0]{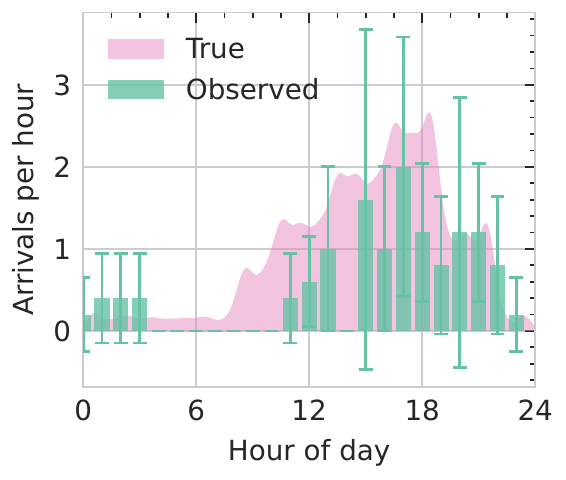}};
\node[emphbox, anchor=north, minimum width=0.32\linewidth](plot3interobs) at ([yshift=-0mm] plot3.south)
    {Interobs. Time of \circnum{3}: \numWithUncertaintyFromCSV[round-mode=places,round-precision=2]{04_media/observations_a6ad4699-53c9-4d01-b8d2-5be8540d6378.csv}{interObservationTime}{\second}};

\end{tikzpicture}
\caption{The local subgraph around place \protect\circnum{1}, located near a school, is shown in (a). Panel (b) depicts an observation heatmap, where small black circles indicate few and large red-to-white circles indicate many agent observations, highlighting uneven coverage along routes. Panels (c–e) report bicycle arrival rates at nodes \protect\circnum{1}~–~\protect\circnum{3}, comparing true distributions obtained from long-run Monte Carlo sampling (pink) with observed beliefs averaged over five replications (green, error bars are standard deviation). While node \protect\circnum{1} is in the vicinity of a school, nodes \protect\circnum{2} and \protect\circnum{3} are located in purely residential areas.}
\label{fig:experiments}
\end{figure*}
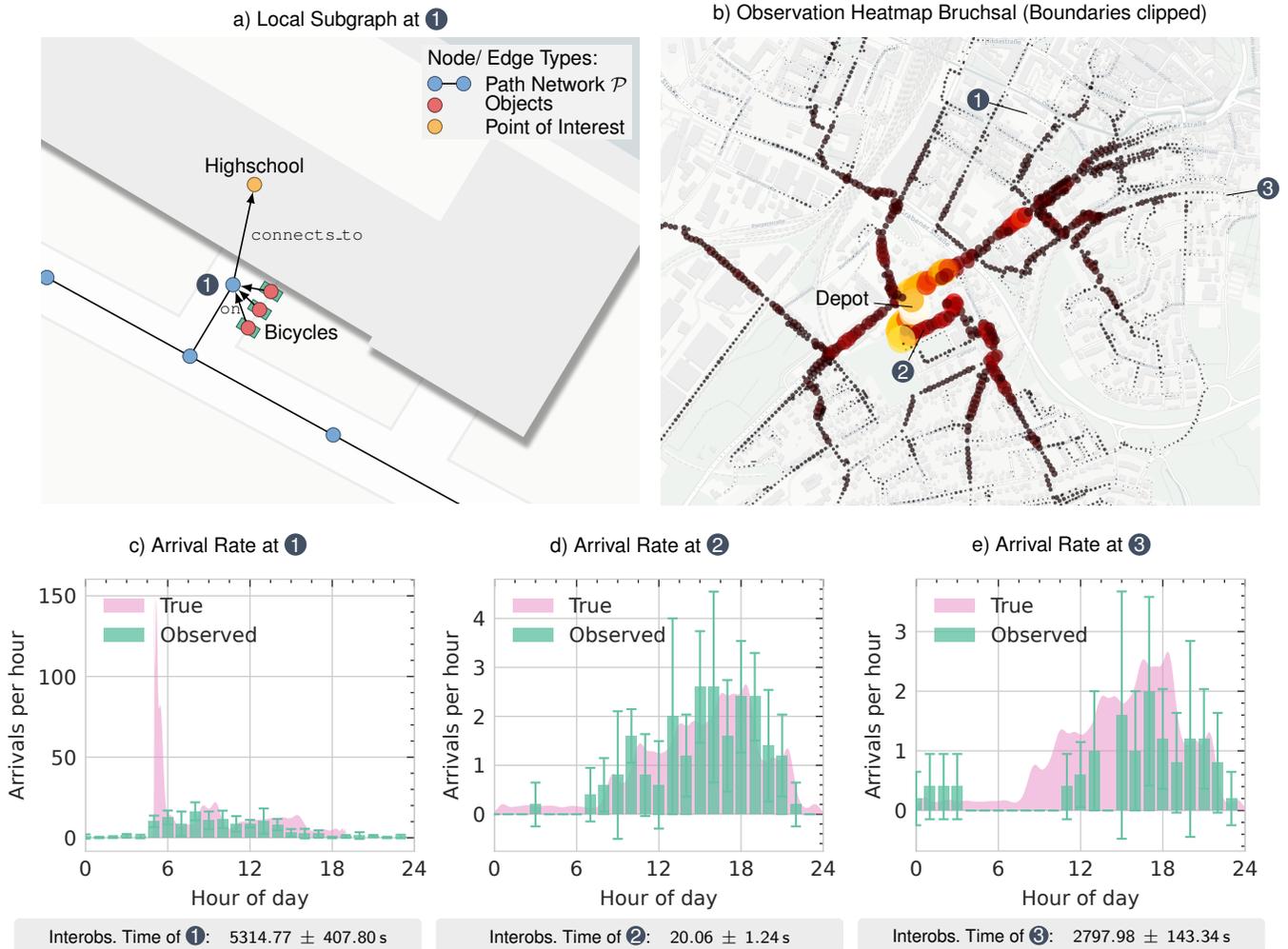

%% file: 01_sections/04_application.tex
\section{Application: Autonomous Delivery Robots}\label{sec:application}
Motivated by the availability of semantic data through \ac{osm} \cite{OpenStreetMap}, limited availability of simulation environments, and possible applications (e.g., controllable urban scene generation \cite{liuControllable3DOutdoor2025}), we demonstrate the concept of \fogmachine in urban environments with \acp{adr} as mobile agents moving on sidewalks. 
First, we describe the modeling and parametrization process, then select three scenarios in different-sized urban areas in Germany. For each scenario, we employ \fogmachine as a Monte Carlo Simulation and derive stochastic distributions of semi-static objects like parked cars and bicycles and measure their impact on the navigation of the mobile agent fleet.
\input{02_figures/fig03_daily_trends}

\subsection{Scene Graph Generation}
The initial \ac{sg} $\mathcal{G}_0$ is automatically constructed from \ac{osm} data within a defined radius, constrained by the robot's range using the L1 distance. First, the pedestrian network is extracted and interpolated with additional interpolated nodes inserted along edges to control spatial resolution. \acp{poi}, such as buildings and amenities, are then retrieved from OpenStreetMap and filtered according to semantic tags. The building closest to the central coordinates is defined as a central hub (micro depot, see \cite{mullerDesignRealWorldApplication2024}). Each \ac{poi} is incorporated into the \ac{sg} and linked to its nearest traversable node via an access edge. To ensure structural consistency, nodes outside the reachable area and \acp{poi} without an access edge are removed. Additionally, only the largest connected subgraph is retained.
Based on available semantic tags, we extract place nodes with six different semantic classes: \texttt{sidewalk}, \texttt{work}, \texttt{education}, \texttt{leisure}, \texttt{housing}, and \texttt{retail}.

\subsection{Modeling and Parametrization}
For this application, we did choose simplistic models to describe different components of the system. These models are parametrized, when available, from real-world statistics\footnote{e.g., \url{https://mobilitaet-in-tabellen.bast.de/login.html?brd}, documentation in paper repository.}, providing a realistic mean behavior of the simulation.

\textbf{Processes}
are characterized through \acp{nhpp} (see \cref{sec:stoch_models}) to provide a realistic, time-variant model with daily periodicity for commonly occurring, semi-static object types:  $\mathcal{C}_{d} := \{\texttt{car}, \texttt{bicycle}, \texttt{trashcan}\}$. We further consider buildings as \acp{poi}, $\mathcal{C}_q := \{\texttt{work}, \texttt{education}, \texttt{leisure}, \texttt{housing}, \texttt{retail}\}$. We then instantiate a process for every combination of \ac{poi} instance and object type. Each process uses an \ac{nhpp} with a hourly-discrete arrival rate for spawning ($q_\pi$) and a exponentially-distributed lifetime function $l_\pi$, conditioned through the above mentioned open-system approximation. We choose Germany as the closed system and use country-wide statistics to calculate interdependency. 

Objects drain only to nodes of the path network. Each node in $\mathcal{P}$ is assigned a capacity for each object class, determined by its semantic category, which constrains the number of objects that can be attached to it. Upon arrival at a \ac{poi}, an object drains to the nearest node on the traversable path network that still has available capacity for its class using Dijkstra's algorithm. This \textbf{capacity-based object passing} simulates human behavior in minimizing walking distance, under the assumption that individuals always identify and choose the closest location to place the object.


Mobile Agents perform delivery \textbf{Tasks} analogously to \textcite{mullerDesignRealWorldApplication2024}, generated by \ac{poi} nodes with interarrival times drawn from an \ac{nhpp}. This enables us to define daytime-dependent arrival rates which replicate real-life task distributions, e.g., increased volumes of food delivery tasks during lunchtime.
To align with the urban delivery scenario, we assume that a delivery task entails the transportation of a good from the depot node to the \ac{poi} $v_\text{PoI}$ which issued the task. Once a delivery task is issued, it is assigned to the next available mobile agent in the depot. The agents are constrained to process only one task at a time. An agent must return to the depot after task completion.

\textbf{Mobile Agents} traverse the path network to accomplish these delivery tasks. To account for the fact that other objects act as obstacles in the constrained sidewalk environment, we model the travel time over a place $v_j$, $w\left(v_j\right)$, using a simple linear density model. By providing parameters for the average sidewalk segment length $l_s$, the sidewalk width $b_s$, and the agent width $b_\text{agent}$, we can calculate the available area $A_\text{free}$ which is not swept by the agent with $A_\text{free}= l_s \cdot (b_s - b_\text{agent})$. Based on the areas $A_i$ each object $o^{(i)}$ contained in place $v_j$ occupies, we linearly decrease the agent's velocity,
\begin{equation}
    \nu_{j} = \text{max} \left( \frac{A_\text{free} - \sum A_i}{A_\text{free}} \cdot \nu_\text{default}, 0 \right),
\end{equation}
from which the \textit{additional} travel cost
\begin{equation}
    w_{penalty,j} = \frac{l_s}{\nu_j} - \frac{l_s}{\nu_\text{default}}
\end{equation}
for passing node $v_j$ is calculated. If $\nu_j = 0$, the agent must wait until enough objects despawn or replan their path.

Path planning is performed dynamically on the network $\mathcal{P}$ using A* \cite{hartFormalBasisHeuristic1968} and employing the observed scene graph $\mathcal{G}_\text{obs}$ to estimate delays through obstacles. It utilizes the dynamic weight function $w = w_{static} + w_{penalty}$,
where $w_{static}$ stands for the static travel cost based on edge distances and the agent's default
velocity and $w_{penalty}$ represents penalties originating from objects
located at edge target nodes in $\mathcal{G}_\text{obs}$. 

\textbf{Observation Model:}
We assume equal sensor setups for all mobile agents, resulting in a uniform observation function $\sigma_i = \sigma, \forall a_i \in \mathcal{A}$.
To model limited sensor range, an observation at $t_i$ updates the neighborhood within a range of $r_\sigma$, thus
\begin{equation}
    \sigma(\mathcal{G}[t_i]) = \mathcal{G}[t_i]\big[\, \{ v \in \mathcal{V}[t] \mid \lVert a_i - v \rVert_2 < r_\sigma \}\,\big],
\end{equation}
where $\mathcal{G}[t][\cdot]$ denotes the induced subgraph on the selected node set. We assume a perfect sensor system and therefore neglect
influence factors originating from sensor noise.

\subsection{Scenarios}\label{sec:scenarios}
We carefully selected three scenarios to provide a comprehensive analysis of \fogmachine's flexibility and diversity. Due to parameter availability, we constrain the scenarios to German cities:

\textbf{Bruchsal} is a mid-sized German city with diverse urban structures, including all defined semantic classes. Bruchsal already hosts a test area for autonomous delivery \cite{mullerDesignRealWorldApplication2024} which provides practical relevance and future opportunities for future real-world validation.

\textbf{Wenningstedt} is a small, low-density settlement characterized by tourism, seasonal housing, and local services. Its morphology contrasts urban settings, making it suitable for testing the adaptability of the \ac{sg} approach. The town is dominated by housing and leisure \acp{poi}, while work, education, and retail classes are nearly absent. This highlights how object distributions and arrival patterns shift when certain \acp{poi} are dominant or missing. In rural areas, more cars and bicycles are parked on private ground; hence, the assignment probability to sidewalk nodes is reduced to 40\% for cars and 20\% for bicycles.

\textbf{Trier’s} city center is characterized by a dense medieval street network, creating a different spatial structure. As a cultural and touristic hub, Trier’s center has many retail and leisure \acp{poi}. The overall density of \ac{poi} is relatively high (see \ref{tab:stats}), providing a scenario that highlights the system’s behavior under conditions with a larger number of arriving objects.

\subsection{Experiments}\label{sec:experiments}
\input{02_figures/table01_statistics}

We evaluate \fogmachine along three central dimensions: (1) the ability to reproduce \textbf{emergent behavior} at the system level, (2) the impact of \textbf{partial observability} on the reconstruction of the true system state and on the task performance of mobile agents, and (3) the \textbf{efficiency} of the underlying \ac{des}.  
All evaluations are conducted across the three representative scenarios introduced in \cref{sec:scenarios}, allowing us to assess both the expressiveness of the modeling approach and the scalability of the simulator.

For each scenario, we conduct simulations of $22$ days with five replications per scenario. Prior experiments showed system stabilization after approximately two days of simulation. Consequently, we omit the recordings of the first $48$ hours from all evaluation metrics. True distributions (\cref{fig:experiments}) are derived from sampling over a total of $40,000$ days.

To study \textbf{emergent behavior}, we analyze temporal processes such as arrival and lifetime distributions per place. These are sampled per hour of day for both the true scene graph $\mathcal{G}$ and the observed scene graph $\mathcal{G}_\text{obs}$. This aggregation highlights the characteristic daily cycles of objects and their heterogeneity across scenarios.

To capture the impact of \textbf{partial observability}, we quantify the mismatch between the true state $\mathcal{G}$ and the observed state $\mathcal{G}_\text{obs}[t]$.  
We employ several observability metrics: the \emph{inter-observation time}, i.e., the interval between two consecutive observations of the same node, with the average inter-observation time per node providing a measure of observation frequency; the \emph{up-to-date node share}, i.e., the percentage of simulation time during which all objects at nodes $v^{(i)}_P \in \mathcal{V}_P$ are correctly reflected in $\mathcal{G}_\text{obs}[t]$; and the \emph{task delay}, defined as the normalized difference between predicted and actual task completion times,
\begin{equation}
    d = \frac{t_{\text{pred}} - t_{\text{true}}}{t_{\text{true}}}.
\end{equation}
Averaging $d$ over all executed tasks yields the average task delay. To assess the role of dynamic scene graphs in navigation, we compare two planners: (i) a baseline planner using only the static path network $\mathcal{S}$, and (ii) a graph-aware planner operating on the observed dynamic scene graph $\mathcal{G}_\text{obs}$.

Finally, to assess \textbf{DES efficiency and effectiveness}, we first characterize the scale of each scenario to contextualize load conditions. We report the path network size ($\lvert\mathcal{V}_P\rvert$), the number of \acp{poi}, the average arrival rate per \ac{poi}, and the average arrival rate per path network node. These structural and dynamic metrics define the conditions under which the simulator operates. We then measure computational efficiency in terms of  the real-time factor (ratio of simulated time to wall-clock time). All experiments were executed on an \SI{4.7}{\giga\hertz} Intel© Core™ i9-10900X CPU.

\subsection{Results}
\label{sec:results}
The outcomes of the experiments are summarized in two figures and one table: daily temporal patterns are shown in \cref{fig:daily_trends}, spatial and distributional effects in \cref{fig:experiments}, and scenario statistics in \cref{tab:stats}. The discussion below follows the three evaluation dimensions introduced in Sec.~\ref{sec:experiments}.

\textbf{Emergent behavior:}  
\cref{fig:daily_trends} shows daily patterns for cars, bicycles, and trashcans. Across all scenarios, cars, and bicycles follow strong daily cycles with maxima at night and minima during late morning to noon, while trashcans remain nearly time-invariant. Relative variation is largest in Wenningstedt, moderate in Bruchsal, and smallest in Trier, consistent with the distribution of different \ac{poi} classes in each scenario as discussed in \cref{sec:scenarios}.

Beyond these global trends, local interactions also give rise to emergent spatial phenomena. Route planning of mobile agents produces heterogeneous observation frequencies, as seen in the Bruchsal heatmap (\cref{fig:experiments}b): nodes close to major routes or the central depot are updated much more frequently than peripheral ones. At the same time, nodes in similar neighborhoods—such as residential areas around nodes~\circnum{2} and~\circnum{3}—show comparable arrival distributions. Ground-truth arrival and lifetime distributions further reveal daily seasonality and concentrated activity at highly frequented areas (e.g., node~\circnum{1} near a school), demonstrating realistic dynamics emerging from simple rules.  

\textbf{Partial observability:}  
Scenarios with lower activity (e.g., Wenningstedt) achieve a higher up-to-date node share, while busier ones (Bruchsal, Trier) show more outdated nodes. Estimating distributions from observations proves difficult due to their partial nature and inherent delays. This is emphasized by comparing nodes~\circnum{2} and~\circnum{3}, which have similar true arrival distributions but diverging observed beliefs, as node~\circnum{2} is visited far more often. The task-delay metric highlights the practical consequences: when planning is based only on the static graph $\mathcal{S}$, delays are significant, but drop to nearly zero once object observations are included in $\mathcal{G}_\text{obs}$. The gap between both planners underscores the importance of modeling semi-static objects.

\textbf{DES efficiency:}  
Across all scenarios, the simulator maintains favorable real-time factors. The RTF scales antiproportionally with scene graph size, yet the decrease is far less pronounced than would be expected for agent-based simulations, demonstrating the scalability of the discrete-event approach.

%% file: 02_figures/fig03_daily_trends.tex
\begin{figure*}
    \centering
    \includegraphics[scale=1]{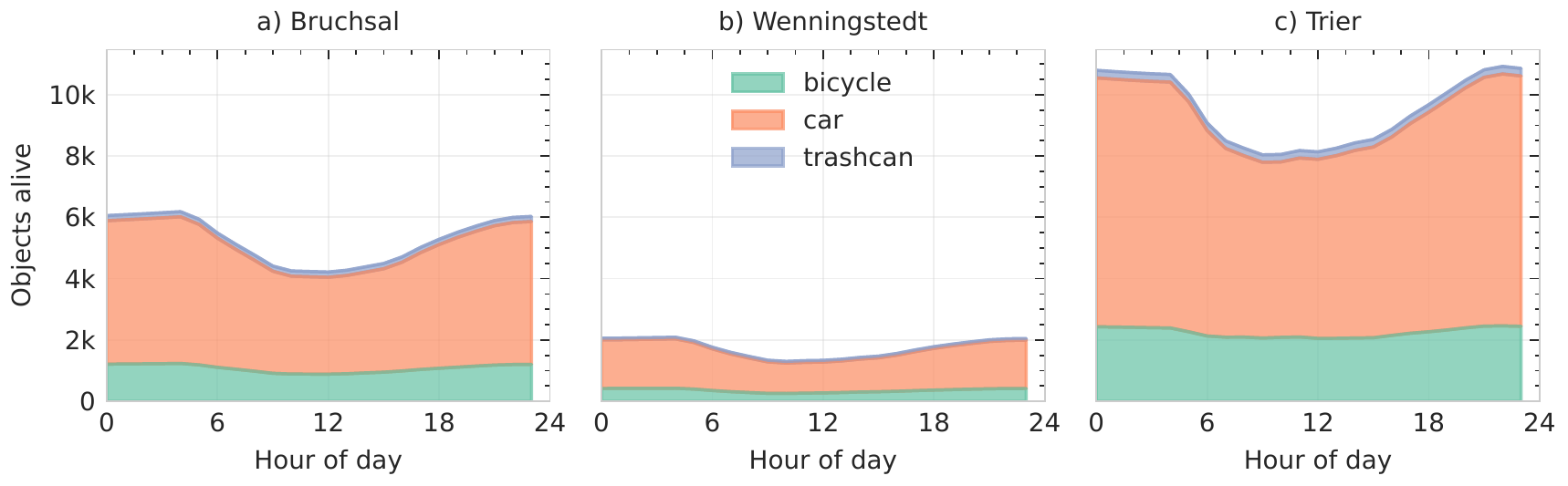}
    \caption{Daily Trends: Number of objects alive in the simulation for a) Bruchsal, b) Wenningstedt, and c) Trier. }
    \label{fig:daily_trends}
\end{figure*}

%% file: 02_figures/table01_statistics.tex
\begin{table*}
\centering
\caption{Statistics and Metrics for each of the Three Scenarios}
\label{tab:stats}
\begin{tabularx}{\linewidth}{X
    c@{} 
    c@{} 
    c@{} 
    c@{} 
    c@{} 
    c@{} 
    c@{} 
    c@{} 
}
\toprule
\multirow{2}{*}{Scene Graph}%
& \multicolumn{1}{c}{\multirow{2}{*}{\parbox{10mm}{\centering \# Nodes $\in \mathcal{P}$ }}} 
& \multicolumn{1}{c}{\multirow{2}{*}{\parbox{10mm}{\centering \# PoIs }}} 
&\multicolumn{2}{c}{{Arrival Rate of Objects  [\si{h^{-1}}] per ...}} 
& \multicolumn{1}{c}{\multirow{2}{*}{\parbox{10mm}{\centering\metricUp  RTF }}} 
& \multicolumn{1}{c}{\multirow{2}{*}{\parbox{22mm}{\centering \metricUp Up-to-date nodes [\si{\percent}]}}} 
&\multicolumn{2}{c}{{\parbox{35mm}{\centering\metricDown Average Task Delay [\si{\percent}]}}} \\

\cmidrule(lr){4-5}\cmidrule(lr){8-9}
&
&
&
{... PoI} 
& {... Node} 
&
&
& {No obj.} 
& {Obs.} \\

\midrule
\begin{filecontents*}{04_media/table01_data.csv}
place,spawnPoiMean,spawnPoiStd,spawnNodeMean,spawnNodeStd,upToDateMean,upToDateStd,delayNoObstMean,delayNoObstStd,delayObsMean,delayObsStd,adrs,pois,nodes, rtf
Bruchsal,4.4755744949494956,0.010504480685864917,2.8242012588638357,0.0066285942978288465,0.7589155961780205,0.0008318703706868604,0.20608244861315014,0.0018451262146993313,0.00388097943888492,2.9096614333406483e-05,70,2880,7445,79.2
Wenningstedt,5.187149902298316,0.010333523641196758,1.8902817808823031,0.003765704064695617,0.8184780836949391,0.0010195651059692798,0.1412220006279742,0.0020255375457311157,0.00518769106436984,5.6655433825509925e-05,17,977,3659,833.6842
Trier,10.709305387441338,0.014187550534024485,6.089326120817687,0.00806706120815021,0.7371783399361016,0.0009208219645893746,0.21318035996251852,0.0012626521509225977,0.00520270268622882,2.206361159874915e-05,86,4281,11811, 48.0
\end{filecontents*}
\csvreader[head to column names]{04_media/table01_data.csv}{}{%
  \place
  & \nodes
  & \pois
  & \num[round-precision=2, round-mode=places]{\spawnPoiMean} $\pm$ \num[round-precision=2, round-mode=places]{\spawnPoiStd}
  & \num[round-precision=2, round-mode=places]{\spawnNodeMean} $\pm$ \num[round-precision=2, round-mode=places]{\spawnNodeStd}
  & \num[round-precision=2, round-mode=places]{\rtf}
  & \numpercent[round-precision=2, round-mode=places]{\upToDateMean} $\pm$ \numpercent[round-precision=2, round-mode=places]{\upToDateStd}
  & \numpercent[round-precision=2, round-mode=places]{\delayNoObstMean} $\pm$ \numpercent[round-precision=2, round-mode=places]{\delayNoObstStd}
  & \numpercent[round-precision=2, round-mode=places]{\delayObsMean} $\pm$ \numpercent[round-precision=2, round-mode=places]{\delayObsStd}\\
}
\\
\bottomrule
\end{tabularx}

\vspace*{0.5em}

\metricUp\, indicates higher is better; \metricDown\, indicates lower is better; numerical values are mean\,$\pm$\,std evaluated on 5 simulation runs; RTF = Real-Time Factor

\end{table*}

%% file: 01_sections/05_conclusion.tex
\section{Conclusion}
\label{sec:conclusion}

The presented results demonstrate that \fogmachine successfully models large, interconnected environments under partial observability while maintaining efficiency through discrete-event simulation. Daily activity patterns, arrival distributions, and the gap between true and observed states highlight both the expressive power of \fogmachine and the challenges of reasoning in such environments. In particular, scenarios with similar ground-truth dynamics but diverging beliefs underline the impact of delayed and partial observations on situational awareness. This shows that observation strategies, rather than the environment alone, critically shape the agents' perception capabilities.

From a computational perspective, the simulator achieves favorable real-time factors. This scalability makes \fogmachine well suited for long-term evaluations and Monte Carlo analysis of stochastic \acp{dsg}, both of which are essential when studying rare events or uncertainty propagation.

At the same time, several limitations remain. The current setup was showcased in an application with large but shallow scene graphs, while effectiveness in deeply hierarchical domains (e.g., household environments) has yet to be shown. Moreover, the estimation of uncertainty within \acp{dsg} emerged as a central challenge in the experiments. More elaborate strategies are required, potentially including active exploration and information-gain–driven planning.

Future work will therefore explore downstream tasks in direct combination with \fogmachine. For model training, more comprehensive process models are needed. Promising directions include real2sim$\rightarrow$sim2real pipelines, where real-world observations parametrize simulation processes that can then be scaled to diverse scenarios, or augmenting processes with world knowledge via LLMs acting as prototype agents for specific PoIs. Another compelling perspective is to employ \fogmachine for temporally consistent conditioning of scene-generative models, thereby enabling realistic generation of environments over extended horizons.

%% file: 01_sections/98_contributions.tex
\section*{CONTRIBUTIONS}
LO: Overall idea \& writing \ref{sec:introduction}-\ref{sec:methods}, \ref{sec:conclusion}; NH \& MW: implementation, experiments, and writing \ref{sec:application}; KF: Overall project supervision; all: proofreading.

%% file: 01_sections/99_acknowledgements.tex
\section*{ACKNOWLEDGMENT}\label{sec:acknowledgment}
Map data used for \cref{fig:experiments} and experiments in \cref{sec:application} is copyrighted by OpenStreetMap contributors and available from \url{https://www.openstreetmap.org}. \fogmachine is implemented using SimPy (\url{https://gitlab.com/team-simpy/simpy}).